\RequirePackage{fix-cm}
\documentclass[smallextended]{svjour3}       % onecolumn (second format)
\smartqed  % flush right qed marks, e.g. at end of proof
\usepackage{graphicx}
\usepackage[table]{xcolor}% http://ctan.org/pkg/xcolor
\usepackage{url}
\usepackage{natbib}

\begin{document}

\title{Quantitative Fine-Grained Human Evaluation of Machine Translation Systems: a Case Study on English to Croatian
}


\author{Filip Klubi\v{c}ka         	\and
        Antonio Toral				\and 
        V\'{i}ctor M. S\'{a}nchez-Cartagena %etc.
}


\institute{F. Klubi\v{c}ka \at
              Dublin Institute of Technology \\
              \email{filip.klubicka@mydit.ie}           %  \\
           \and
           A. Toral \at
              University of Groningen \\
              \email{a.toral.ruiz@rug.nl}           %  \\
			\and
           V. M. S\'{a}nchez-Cartagena \at
              Prompsit Language Engineering \\
              \email{vmsanchez@prompsit.com}           %  \\
}

\date{Received: date / Accepted: date}

\maketitle

\begin{abstract}
%Insert your abstract here. Include keywords, PACS and mathematical subject classification numbers as needed.

This paper presents a quantitative fine-grained manual evaluation approach to comparing the performance of different machine translation (MT) systems.
We build upon the well-established Multidimensional Quality Metrics (MQM) error taxonomy and implement a novel method that assesses whether the differences in performance for MQM error types between different MT systems are statistically significant.
We conduct a case study for English-to-Croatian, a language direction that involves translating into a morphologically rich language, for which we compare three MT systems belonging to different paradigms: pure phrase-based, factored phrase-based and neural.
First, we design an MQM-compliant error taxonomy tailored to the relevant linguistic phenomena of Slavic languages, which made the annotation process feasible and accurate.
Errors in MT outputs were then annotated by two annotators following this taxonomy.
Subsequently, we carried out a statistical analysis which showed that the best-performing system (neural) reduces the errors produced by the worst system (pure phrase-based) by more than half (54\%).
Moreover, we conducted an additional analysis of agreement errors in which we distinguished between short (phrase-level) and long distance (sentence-level) errors.
We discovered that phrase-based MT approaches are of limited use for long distance agreement phenomena, for which neural MT was found to be especially effective.

\keywords{neural machine translation \and statistical machine translation \and phrase-based machine translation \and factored models \and human evaluation \and error annotation \and multidimensional quality metrics (MQM)}
\end{abstract}

\section{Introduction}\label{s:intro}

A machine translation (MT) paradigm based on deep neural networks, usually referred to as neural MT (NMT) \citep{Bahdanau2014}, has emerged in the past few years.
This has disrupted the MT field since NMT, despite its infancy, has already surpassed the performance of phrase-based MT (PBMT) \citep{koehn2003statistical}, the mainstream approach to date.

The vast potential of NMT in terms of overall performance scores, be those automatic (e.g. BLEU) or human (e.g. system rankings) was, for example, showcased in 2016 news translation shared task at WMT,\footnote{\url{http://www.statmt.org/wmt16/translation-task.html}} where NMT systems significantly outperformed PBMT in 8 of the 9 language directions submitted where NMT systems were submitted, according to human evaluations (system rankings). In these evaluations, users (mainly MT researchers) were presented with a source-language sentence, its reference translation and a set of machine translations produced by the different systems submitted to the shared task. They had to rank the machine translations. 

Additionally, monolingual direct assessment adequacy and fluency evaluations were also carried out in WMT 2016 for translations directions into English. In these evaluations, users had only to give an adequacy and fluency score to individual translations. Whereas the language pairs for which NMT outperformed PBMT according to the adequacy evaluation completely matched those in the system ranking (the only language pair in which NMT did not outperform PBMT was Russian-to-English), the fluency direct assessment showed that NMT output is more fluent than PBMT output for all the language pairs evaluated (including Russian-to-English).

In 2017 edition of the same shared task,\footnote{\url{http://www.statmt.org/wmt17/translation-task.html}} the trend has gained strength and, for all language directions, the best-performing submitted system either follows the NMT architecture or is a hybrid system that includes an NMT component.

The fine-grained human evaluation presented in this paper greatly differs from WMT evaluation: instead of just ranking translations, the annotators had to classify the errors contained in each translation produced by the MT systems being evaluated according to a complete error hierarchy and choose the particular tokens that contains the error.

Considering the high overall performance of NMT, researchers have in the past year attempted to analyse the potential of NMT in more detail.
While overall scores, such as those obtained in WMT evaluation, give an indication of the general performance of a system, they do not shed light on the strengths and weaknesses of this new paradigm to MT. Hence, two recent papers have looked at automatically conducting multifaceted evaluations:
\begin{itemize}
\item 
\citet{D16-1025} performed a detailed analysis of the English-to-German language direction, comparing state-of-the-art PBMT and NMT systems on transcribed speeches.
Their findings show that NMT (i) decreases post-editing effort, (ii) degrades faster than PBMT with sentence length and (iii) improves notably on reordering and inflection.
\item
\citet{DBLP:journals/corr/ToralS17} carried out a series of analyses and evaluations for NMT and PBMT systems on the news domain for 9 language pairs.
Their research corroborated the findings of \citet{D16-1025} regarding NMT's excellent performance on reordering and inflection and its degradation with sentence length.
In addition to that, \citeauthor{DBLP:journals/corr/ToralS17}'s findings show that NMT systems (i) exhibit higher inter-system variability, (ii) lead to more fluent outputs and (iii) perform more reordering than PBMT, but less than hierarchical PBMT.
\end{itemize}

A limitation of these analyses lies in the fact that all of them were performed automatically (e.g. reordering and inflection errors were detected based on automatic evaluation metrics).
More recently, other authors have performed human analyses of NMT's strengths and weaknesses in comparison with PBMT and rule-based paradigms. Such human evaluations do not suffer from the potential biases introduced by automatic tools employed in the above papers.
\begin{itemize}
\item \citet{aljoscha2017} presented a study based on an error categorization specifically tailored to the English--German language pair (in both directions) and a test set carefully designed in order to cover the most relevant linguistic phenomena. They conclude that NMT systems are able to produce translations that resemble those produced by rule-based MT without using explicit linguistic information.
\item \citet{popovic2017} also targeted the English--German language pair and identified language-related issues in the outputs of NMT and PBMT systems. She concluded that NMT systems are better than PBMT ones in handling verbs, English noun collocations, German compound words, phrase structure and articles, while PBMT systems perform better when dealing with prepositions, translation of English (source) ambiguous words and generation of English (target) continuous tenses. As the issues are complementary between the two MT paradigms analysed, results suggest that hybridisation between them could be a promising way forward.
\item \citet{castilho2017} evaluated the performance of NMT versus PBMT for three different translation domains: e-commerce product listings, patents and massive open online courses. They performed error analysis with an error taxonomy consisting of 7 categories for patent translation from Chinese to English. The analysis showed that NMT made more omission errors than PBMT, while PBMT systems made more errors related to sentence structure than NMT.
Overall, they concluded that, according to human evaluation, NMT has not fully reached the quality of PBMT.

\end{itemize}

This paper adds to the body of research dealing with manual analysis of NMT systems by conducting a detailed human analysis of the outputs produced by NMT and PBMT systems when translating news texts in the English-to-Croatian language direction. We manually annotate the errors found according to a detailed error taxonomy that is compliant with the hierarchical listing of issue types defined as part of the Multidimensional Quality Metrics (MQM)~\citep{mqm}. 
First, we define an error taxonomy that is relevant to the problematic linguistic phenomena of this language pair.
Subsequently, we annotate the errors produced by 3 state-of-the-art translation systems that belong to the following paradigms: PBMT, factored PBMT \citep{koehn2007} and NMT.
Finally, we analyse the annotations and draw conclusions.

This paper's main contribution can thus be summarised as follows:
\begin{enumerate}
\item We conduct one of the first human fine-grained error analyses of NMT in the literature and, to the best of our knowledge, the first one in which a Slavic language is involved.
\item We analyse NMT in comparison not only to pure PBMT and hierarchical PBMT, as in other previous work, but also with respect to factored models.
\item We develop an MQM-compliant error taxonomy for Slavic languages. It is much more detailed in terms of error categories than that followed by \citet{castilho2017} in their Chinese-to-English human evaluation, to account for the grammatical features of Slavic languages. Additionally, unlike the taxonomies used by \citet{aljoscha2017} and \citet{popovic2017}, ours is not restricted to a single language pair, and is at the same time based on a well-known error categorization framework (MQM).
\item Unlike \citet{aljoscha2017} and \citet{popovic2017}, we included two annotators in our evaluation so that each sentence is annotated twice. This allows us to compute inter-annotator agreement, which increases the reliability of our results.
\item We also employ a statistically grounded approach to analyzing and interpreting the results of MQM error annotation that goes beyond simple counting of errors.

\end{enumerate}

This paper builds upon our recent work on this topic~\citep{filip2017}, which is here extended in a number of directions:

\begin{enumerate}
\item We have performed additional categorisation and analysis of agreement errors, in order to investigate whether there is a difference in the number of agreement errors produced in regards to their scope, i.e. we looked at whether the reduction in agreement errors equally affect phrase (or short distance) agreement and sentence (or long distance) agreement.
\item We have included some examples of sentences from the dataset used in the experiments to better illustrate the different MQM error types.
\item We have included a more detailed discussion, expanded some points and added an explanation of the statistics calculated from the MQM annotation.
\end{enumerate}

The rest of the paper is organized as follows.
Section \ref{s:sys} describes the MT systems and the datasets used in our experiments.
Section \ref{s:mqm} includes the definition of the error taxonomy and explains the annotation setup and guidelines given to annotators. Next, Section~\ref{s:results} presents the results obtained and their discussion.
Section \ref{s:additionalagreement} describes the additional annotation focused on agreement errors and analysis thereof.
Finally, Section \ref{s:con} outlines the conclusions and lines of future work.

\section{MT Systems and Datasets}\label{s:sys}

This section describes the MT systems and the datasets used in our experiments. We built PBMT, factored PBMT and NMT systems.

The 3 systems were trained on the same parallel data.
We considered a set of publicly available English--Croatian parallel corpora, comprising the DGT Translation Memory,\footnote{\url{https://ec.europa.eu/jrc/en/language-technologies/dgt-translation-memory}} HrEnWaC,\footnote{\url{https://www.clarin.si/repository/xmlui/handle/11356/1058}} JRC Acquis,\footnote{\url{http://tinyurl.com/CroatianAcquis}} OpenSubtitles 2013,\footnote{\url{http://www.opensubtitles.org/}} \textsc{SETimes}\footnote{\url{http://opus.nlpl.eu/SETIMES2.php}} and \textsc{Ted} talks\footnote{\url{http://opus.nlpl.eu/TedTalks.php}}, many of which can be obtained from OPUS\footnote{\url{http://opus.nlpl.eu}} \citep{Tiedemann:RANLP5,TIEDEMANN12.463}.  
We concatenated all of these corpora and performed cross-entropy based data selection~\citep{Moore:2010:ISL:1858842.1858883} using the development set.
Once the data is ranked we keep the 25\% highest-ranked sentence pairs (4,786,516). Data selection was carried out in order to speed up training and discard the training parallel sentences that are too different from the domain of the development and test sets (news) and hence could have a negative impact on the results.

PBMT systems also require monolingual data for language modeling.
To this end we concatenated the hrWaC corpus~\citep{ljubesic14-bs} with the target side of the aforementioned parallel corpora.

As our development set we used the first 1,000 sentences of the English test set used at the WMT12 news translation task,\footnote{\url{http://www.statmt.org/wmt12/translation-task.html}} translated by a professional translator into Croatian.
Similarly, our test set is comprised of the first 1,000  sentences of the English test set of the WMT13 translation task,\footnote{\url{http://www.statmt.org/wmt13/translation-task.html}} again manually translated into Croatian.

The PBMT system was built with Moses v3.0\footnote{\url{https://github.com/moses-smt/mosesdecoder/tree/RELEASE-3.0}} \citep{koehn2007moses}.
In addition to the default models we also used
hierarchical reordering~\citep{galley2008simple}, an operation sequence model~\citep{durrani2011joint} and a bilingual neural language model~\citep{devlin2014binlm}.

The factored PBMT system maps one factor in the source language (surface form) to two factors in the target (surface form and morphosyntactic description).
This system is described in detail by~\citet{sanchez2016dealing}.

The NMT system is based on the sequence-to-sequence architecture with attention~\citep{Bahdanau2014} and it was built with Nematus~\citep{SennrichNematus2017}. We applied sub-word segmentation with byte pair encoding~\citep{sennrich2015a} jointly on the source and target languages.
We performed $85\,000$ join operations.
We defined a hidden layer size of $1\,000$ and an 
embedding layer size of $620$.
We used Adadelta~\citep{zeiler2012adadelta} with
 a minibatch size of $80$, and reshuffled the training set between epochs.
We applied gradient clipping~\citep{icml2013_pascanu13} with a cutoff of $1.0$. 
Training was run for $10$ days and a model was saved every $4.5$ hours.
We decoded the test set using an ensemble of 4 models. These were the 4 models with the highest BLEU scores on the development set.

Table \ref{t:autoeval} reports the scores obtained in terms of the BLEU \citep{papineni2002bleu} and TER \citep{snover2006study} automatic evaluation metrics on the 3 systems previously described.
It can be observed from the table that the use of factored models leads to a substantial improvement upon pure PBMT (6\% relative in terms of BLEU).
NMT allows us to obtain a further notable improvement; 14\% relative in terms of BLEU compared to the factored PBMT system and 21\% compared to the initial PBMT system. All the differences are statistically significant according to paired bootstrap resampling~\citep{koehn2004statistical} ($p\le0.05$, $1\,000$ iterations).

\begin{table}[htbp]
\begin{center}
\begin{tabular}{lrr}
\hline
\bf System & \bf BLEU & \bf TER\\
\hline 
PBMT & 0.2544 & 0.6081\\
Factored PBMT &0.2700 &0.5963\\
NMT &0.3085 & 0.5552\\
%\hline
\hline 
\end{tabular}
\caption{Automatic evaluation (BLEU and TER scores) of the 3 MT systems
\label{t:autoeval}}
\end{center}
\end{table}

\section{Error analysis}\label{s:mqm}

The fact that Croatian is rich in inflection, has rather free word order and other similar phenomena not present in English gives rise to specific translation issues. For example, grammatical categories that do not exist in English, like gender or case inflections in nouns, may be particularly hard to generate reliably in a Croatian translation. We built our factored PBMT system (cf. Section \ref{s:sys}) aiming to directly address such issues. Similarly motivated was our goal to find out how an NMT system would grapple with the same issues. Existing research on this tells us that both systems should lead to improvements on such linguistic aspects. However, this would happen for different reasons: factored SMT deals with explicit linguistic knowledge about grammatical categories, while NMT combined with sub-word representation (e.g. byte pair encoding) solves the problem implicitly in an unsupervised manner, without actually knowing what the grammatical categories are.

Indeed, as shown in Section \ref{s:sys}, both systems lead to significant improvements compared to the pure PBMT system in terms of automatic evaluation metrics.
However, as is the nature of automatic scoring methods, these provide solely an overall score for each system, but do not indicate whether any of the linguistic problems mentioned earlier have been addressed by the systems.
Hence, the question of whether the linguistic quality (or rather, grammaticality) of the output is improved has not been answered by automatic evaluation. Are cases and gender handled better? Has agreement been improved? 

In order to provide answers to these research questions, we decided to thoroughly compare these systems by systematically analyzing their outputs via manual error analysis. In this way we can obtain a more complete picture of what is happening in the translation, which can provide pointers on where to act to obtain further improvements in the future. In the remainder of this section, we describe the annotation framework, overall annotation process and show the level of agreement between the annotators who took part in the process.

\subsection{Multidimensional Quality Metrics and the Slavic tagset}

We decided to make use of the MQM framework, developed in the QTLaunchpad project,\footnote{\url{http://www.qt21.eu/mqm-definition/definition-2015-06-16.html}} for performing the task of manual evaluation via error analysis. It is a framework for describing and defining custom translation quality metrics. It provides a flexible vocabulary of quality issue types and a mechanism for applying them to generate quality scores. It does not impose a single metric for all uses, but rather provides a comprehensive catalogue of quality issue types, with standardized names and definitions, that can be used to describe particular metrics for specific tasks.

The main reason we chose the MQM framework was the flexibility of the issue types and their granularity; it gave us a reliable methodology for quality assessment, that still allowed us to choose which error tags we wanted to use. 

\begin{figure}[h!]
	\centering
	\includegraphics[width=0.9\textwidth]{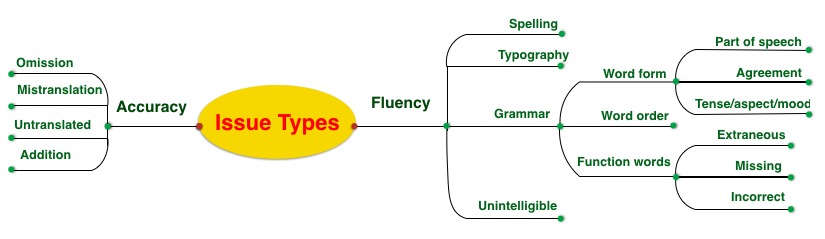}
	\caption{The core set of error categories proposed by the MQM guidelines
    \label{fig:core_set}}
\end{figure}

The MQM guidelines propose a great variety of tags on several annotation layers.\footnote{\url{http://www.qt21.eu/mqm-definition/issues-list-2015-12-30.html}} However, the full tagset is too comprehensive to be viable for any annotation task, so the process begins with choosing the tags to use in accordance with our research questions. 
It is good practice to start with the so-called core tagset, a default set of evaluation metrics (i.e.\ error categories) proposed by the MQM guidelines, shown in Figure \ref{fig:core_set}.

However, given the morphological complexity of Croatian and the way our MT systems were constructed, we found that these core categories were not detailed enough, or rather, did not allow us to conduct an analysis of the specific phenomena we were interested in. Some categories that were of interest to us, like specific \textit{Agreement} types, were not present in the tagset, while some errors, such as \textit{Typography}, were irrelevant to our research questions.

\begin{figure}[h!]
	\centering
	\includegraphics[width=0.7\textwidth]{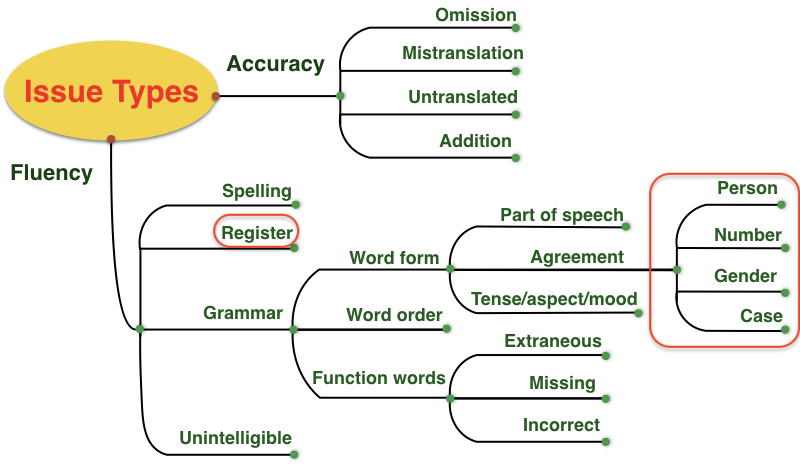}%{slavic_set}
	\caption{The Slavic tagset, a modified version of the MQM core tagset. The additional categories are highlighted with a red rectangle.
    \label{fig:slavic_set}}
\end{figure}

For these reasons, we defined our own set of tags by modifying the core set, rearranging the hierarchy, adding new tags and removing those that were of little relevance.
We call this new tagset ``the Slavic tagset", as its expansion allows for the identification of grammatical errors which are commonly shared by Slavic languages. This tagset is outlined in Figure \ref{fig:slavic_set}.

As evidenced by a comparison of the two figures, we did not change anything about the \textit{Accuracy} branch, but rather modified \textit{Fluency}. As mentioned earlier, we removed \textit{Typography}, but added \textit{Register} in its place. \textit{Register} was included because preliminary insights into the data showed a potential usefulness for annotating a breach of standardness, which has indeed cropped up a couple of times in the systems' outputs. For example, sometimes a synonym for a word can be used, one that is a correct translation in a very general sense, but is actually sub-standard and would not normally be found in that sentence or that particular context (e.g. ``She was the first \textbf{woman} in space." should be translated as ``Bila je prva \textbf{\v{z}ena} u svemiru.", but is instead translated as ``Bila je prva \textbf{\v{z}enska} u svemiru.", roughly corresponding to ``She was the first \textbf{broad} in space." [broad, n. = woman, informal])

In addition to this change, and much more importantly, we added another level to the hierarchy, specifically to the \textit{Agreement} error tag, which we expanded to cover the specific grammatical categories that need to agree in Croatian (nominal categories such as \textit{Gender}, \textit{Number} and \textit{Case}, and the verbal category of \textit{Person}). For example, if the sentence ``The cats walk.", which should be translated as ``Ma\v{c}k\textbf{e} hodaju." is instead translated as ``Ma\v{c}k\textbf{a} hodaju." [The cat walk.], this is to be marked as an error in \textit{Agreement\_Number}.

Given the notoriously low agreement on similar annotation tasks (cf. Subsection \ref{sub:annotatoragreement}), it stands to reason that even the development of such a taxonomy is already prone to human error or disagreement. This is why we made sure that the categories we added were in line with the MQM guidelines; they were already present in the expanded tagset (e.g. \textit{Register}), and those that were not (e.g. the different agreement types) are analogous to tags that are. Still, in order to make sure that we did not taking any missteps in the construction of the taxonomy, we additionally discussed our changes with other researches and colleagues not directly involved in this particular piece of research. Consequently, the taxonomy was verified by both a traditional and computational linguist who respectively specialise in both English and Croatian linguistics.

\subsection{Accuracy versus Fluency}

Unrelated to our interventions in the taxonomy, one important thing to note about the annotation process, as stated in the MQM usage guidelines, is that

\begin{quotation}
``\textit{Accuracy} addresses the extent to which the target text accurately renders the meaning of the source text, whereas \textit{Fluency}, on the other hand, relates to the monolingual qualities of the source or target text, relative to agreed-upon specifications, but independent of relationship between source and target."\footnote{\url{http://www.qt21.eu/downloads/MQM-usage-guidelines.pdf}}
\end{quotation} 
In other words, fluency issues can be assessed without regard to whether the text is a translation or not. So for example, if a translated text tells the user to push a button when the source tells the user not to push it, there is an accuracy issue, while a spelling error or a problem with register remain issues regardless of whether the text is translated.

It has to be said that at first look this distinction might seem obvious and clear-cut, but in practice it is anything but. Very often examples can seem like they belong to either category, and so it is up to the annotators' judgement to decide which level is a better fit, and then being consistent in following through on the decisions made regarding dubious examples.

An example of an error category that might cause trouble for annotators is \textit{Mistranslation}, which describes issues that arise when the content on the target side of the translation does not accurately represent the content on the source side.
The issue is that it can seemingly overlap with the \textit{Fluency} branch;
according to the guidelines, only one error should be tagged, and \textit{Accuracy} trumps \textit{Fluency} if the required information is present in the source text.

\begin{table}[htbp]
\begin{center}
\begin{tabular}{ll}
Source: & For example, \textbf{websites} provide... \\
Correct: & Na primjer, \textbf{internetske stranice} pru\v{z}aju... \\
Translation: & Na primjer, \textbf{internetska stranica} pru\v{z}aju... \\ 
Gloss: & For example, \textbf{website} provide... \\
\end{tabular}
\caption{Example of a \textit{Mistranslation} error that also causes an \textit{Agreement} error.\label{e:mis}}
\end{center}
\end{table}

An example of this is shown in Table \ref{e:mis}, where the only actual error is the translation of `website' in the singular rather than the plural, which is explicitly encoded via the -s morpheme 
in the source text. However, this error then causes a subject-verb agreement error, where the translated subject is singular, but the verb has been correctly translated as plural. This example should, according to the guidelines, be classified only as \textit{Mistranslation}, even though it also shows problems with agreement. If the subject had been translated properly (as the plural), the subject-verb agreement problem would be resolved, so in this case only `internetska stranica' should be tagged as a \textit{Mistranslation}.

\subsection{Annotation setup}

In order to carry out the annotations we used \texttt{translate5},\footnote{\url{http://www.translate5.net/}} a web-based tool that implements annotations of MT outputs using hierarchical taxonomies, as is the case of MQM.

We had two annotators with very similar backgrounds at our disposal. Both are native speakers of Croatian, and both have prior experience with MQM as well as the same academic background; an MA in English linguistics and information science. All of these aspects of the annotators' backgrounds are relevant: their language and linguistics background is necessary given that English is the source language, and Croatian is the target language of our systems, while the information science background promises, at the very least, a basic understanding of what MT is and how it works. Thus, both annotators are well-equipped to handle the task.

Prior to annotation, they were thoroughly familiarized with the \texttt{translate5} system and the official MQM annotation guidelines, which offer detailed instructions for annotation within the MQM framework.\footnote{The instructions include a handy decision tree to aid in the annotation process. It can be found at the following URL: \url{http://www.qt21.eu/downloads/annotatorsGuidelines-2014-06-11.pdf}}

The annotators annotated 100 randomly selected sentences from the test set introduced in Section \ref{s:sys}, while presented with the English source text, a Croatian reference translation and the three unannotated system outputs at the same time. They could choose in which order to annotate, but did not know which translations belonged to which system, thus performing blind annotation. The two annotators did not operate completely independently of each other; they occasionally discussed particularly difficult or ambiguous sentences and how to approach them.

All three translations were annotated by both annotators, meaning that each system translated the same 100 sentences, each annotator annotated the resulting 300 translated sentences (100 source sentences for 3 MT systems), producing a total of 600 annotated sentences (300 translated sentences for 2 annotators). We have made the annotated dataset publicly available on GitHub.\footnote{\url{https://github.com/GreenParachute/mqm-eng-cro/}}

Once the sentences were annotated and the annotation data was extracted, we calculated inter-annotator agreement (reported in Section~\ref{sub:annotatoragreement}) and analyzed the output to determine the performance of each system for each error category (cf. Section~\ref{s:results}).

\subsection{Inter-Annotator Agreement}
\label{sub:annotatoragreement}

Though carefully thought out and developed, the MQM metrics (and manual MT evaluation in general) are notorious for resulting in low inter-annotator agreement (IAA) scores. This is attested by the body of work that has addressed this issue, most notably \citet{lommel-iaa}, who worked specifically on MQM, and \cite{callison-iaa}, who investigated several tasks. This is why it is important that we check how well our annotators agree on the task at hand, and whether this is consistent with prior work done with MQM.

Once the data was annotated, agreement was observed at the sentence level, and inter-annotator agreement was calculated using Cohen's Kappa ($\kappa$) \citep{kappa}.
Agreement was calculated on the annotations of each system separately, as well as on the concatenation of the annotations for the 3 systems together. This way we can (i) investigate whether there are differences in agreement across systems, and also (ii) gain insight into the overall agreement between the two annotators. In addition, Cohen's $\kappa$ was also calculated for every error type separately. Results can be found in Table \ref{t:IAA}.

\begin{table}[!ht]
\begin{center}
\begin{tabular}{lrrrr}
\hline
\bf Error type & \bf PBMT & \bf Factored & \bf NMT & \bf Concat\\
\hline
Accuracy & 0.66 & 0.62 & 0.56 & 0.61 \\
\hspace*{1em}Mistranslation & 0.51 & 0.48 & {\bf 0.58} & 0.53 \\
\hspace*{1em}Omission & 0.34 & 0.39 & 0.37 & 0.37 \\
\hspace*{1em}Addition & 0.50 & 0.54 & 0.33 & 0.47 \\
\hspace*{1em}Untranslated & {\bf 0.86} & {\bf 0.86} & -0.02 & \bf 0.72 \\
Fluency & 0.50 & 0.41 & 0.29 & 0.43 \\
\hspace*{1em}Unintelligible & 0.39 & 0.32 & 0.00 & 0.35 \\
\hspace*{1em}Register & 0.37 & 0.20 & 0.22 & 0.27 \\
\hspace*{1em}Spelling & 0.00 & 0.00 & 0.00 & 0.00  \\
\hspace*{1em}Grammar & 0.50 & 0.43 & 0.33 & 0.45 \\
\hspace*{2em}Word order & 0.56 & 0.33 & 0.21 & 0.40 \\
\hspace*{2em}Function words & 0.43 & 0.27 & 0.36 & 0.35 \\
\hspace*{3em}Extraneous & 0.56 & 0.32 & 0.49 & 0.46 \\
\hspace*{3em}Incorrect & 0.37 & 0.18 & 0.34 & 0.29 \\
\hspace*{3em}Missing & 0.00 & 0.49 & 0.00 & 0.33 \\
\hspace*{2em}Word form & 0.48 & 0.46 & 0.36 & 0.47 \\
\hspace*{3em}Part of speech & -0.03 & 0.10 & 0.00 & 0.04 \\
\hspace*{3em}Tense... & 0.44 & 0.36 & 0.15 & 0.38 \\
\hspace*{3em}Agreement & 0.52 & 0.52 & 0.49 & 0.53 \\
\hspace*{4em}Number & 0.53 & 0.55 & 0.52 & 0.54 \\
\hspace*{4em}Gender & 0.46 & 0.59 & 0.48 & 0.53 \\
\hspace*{4em}Case & 0.53 & 0.49 & 0.52 & 0.56 \\
\hline
\bf All errors & \bf 0.56 & \bf 0.49 & \bf 0.44 & \bf 0.51 \\
\hline 
\bf Any errors & \bf 0.80 & \bf 0.67 & \bf 0.51 & \bf 0.64 \\
\end{tabular}
\caption{Inter-annotator agreement (Cohen's $\kappa$ values) for the MQM evaluation task. The highest score for any individual system and the concatenation, as well as the overall score, are shown in bold. Some of the error categories have no kappa scores attached because they are parent categories that were never used on their own, so there were no data points to calculate the scores.
\label{t:IAA}}
\end{center}
\end{table}

The 'Any errors' IAA value presented at the bottom of the table is the most general agreement measure - it represents agreement on there being any sort of error in a given sentence. These values will logically be higher than the IAA values of the 'All errors' measure (which looks at the total of error agreement, but of specific error categories in a given sentence), and much higher than the agreement calculated for each of the individual, specific error categories. 

Examining the table reveals that our annotators agree most on evaluations of the PBMT system, less so on evaluations of the Factored SMT system, and least on evaluations of the NMT system. The drop in agreement scores for the NMT system is a bit striking. Our intuition is that, because the outputs of the NMT system are much more fluent and grammatically correct (cf. Section \ref{s:results}, errors become less clear cut, and more difficult for our annotators to detect. Or rather, any errors produced by the system are more debatable and the tags are subject to the annotators' interpretation, rather than grounded in some sort of objective truth.

Still, the comparison of IAA between the different systems is likely not that meaningful, as involves a slightly different sample size due to the different lengths of the outputs. Besides, even disregarding this discrepancy, agreement scores are relatively low overall, with the average total $\kappa$ being 0.51. Indeed, the $\kappa$ scores are relatively consistent across all error types for each system, mostly ranging between 0.35 and 0.55. 
According to Cohen, such scores constitute moderate agreement. As already stated, this is to be expected, given the complexity of the problem and annotation schema. In fact, the IAA scores in this work are notably higher than those that have been reported in similar work, e.g. \citet{lommel-iaa}, who achieved $\kappa$ scores ranging between 0.25 and 0.34.

That said, this comparison should be taken with a grain of salt, given that in our setup we looked at sentence-level agreement, while they calculated agreement on the token level. 
The calculations are approached differently here in order to attempt to account for some of the problems that come with span-level annotation. As \citet{lommel-iaa} point out, a ``fundamental issue that the QTLaunchPad annotation encountered was disagreement about the precise scope of errors". In other words, though annotators can agree that a sentence contains the same issue, they might disagree on the span that the issue covers. An example is shown in Table \ref{t:span-agreement} (annotations marked in bold).

\begin{table}[htbp]
\begin{center}
\begin{tabular}{ll}
Source: & Trakhtenberg was the presenter of many programs before Hali-Gali times. \\
Annotator\_1: & Bio \textbf{je} voditelj \textbf{Trakhtenberg} brojnih programa Hali-Gali prije puta. \\
Annotator\_2: & \textbf{Bio je voditelj Trakhtenberg} brojnih programa Hali-Gali prije puta. \\

\end{tabular}
\caption{Example of annotator disagreement on error span on the example of a \textit{Word order} error.\label{t:span-agreement}}
\end{center}
\end{table}

This case shows that annotators can agree on the nature and categorization of issues, yet still disagree on their precise span-level location. Even though they are instructed to mark minimal spans, i.e. spans that cover only the issue in question, they frequently disagree as to what the scope of these issues is. \citet[4]{lommel-iaa} hypothesize that this may be due to the fact that the two reviewers perceive the issue differently, and so see different spans as cognitively relevant. In some instances this disagreement may reflect differing ideas about optimal solutions, while in others the problem may have more to do with perceptual units in the text.

In cases where annotators disagree on the span of the annotation, even \citeauthor{lommel-iaa} are uncertain as to how best to assess IAA. Thus, building on their work and exploring a sentence-level approach is a direction we deemed worth pursuing, as there seems to be no optimal solution, given that both the sentence- and token-level approach come with certain drawbacks. However, to dispel any doubt regarding the reliability of the annotators' judgements on the task at hand, further analysis of the results shows that both annotators' annotations point to comparable conclusions, both when considered separately and together. This is elaborated on in Section \ref{s:results}.

\section{Results
\label{s:results}}

Directly extracting raw annotation data from the \texttt{translate5} system provides a sum of error tags annotated for each error type by each annotator and system. The total values are presented in Table \ref{t:MQM-raw}. 

\begin{table}[!ht]
\begin{center}
\begin{tabular}{lrrr|rrr}
\hline
  & \multicolumn{3}{c|}{\bf Annotator 1}  & \multicolumn{3}{c}{\bf Annotator 2} \\
\hline
System  & \bf PBMT & \bf Factored & \bf NMT & \bf PBMT & \bf Factored & \bf NMT \\
\hline
Total errors & 317 & 276 & 178 & 264 & 199 & 132 \\
\hline
\end{tabular}
\caption{Total errors per system and annotator, ass annotated in MQM.
\label{t:MQM-raw}}
\end{center}
\end{table}

Looking at the aggregate data alone, one can easily detect that both annotators have judged that the PBMT system contains the most errors, and that the NMT system contains the smallest number of errors. This trend is consistent across most fine-grained error categories too, as we will see later on in this section.

However, even though simply counting the errors can provide insight into which system performs better, it does not allow us to draw statistically meaningful conclusions from the results. Error counts cannot be directly compared because different MT systems may output sentences of different lengths, which is indeed the case in the data explored here: in the 100 annotated sentences, the phrase-based system produced an average of 18.99 tokens per sentence, the factored system averaged on 18.89, while the neural system produced 18.36 tokens per sentence. Hence, we need to normalize the scores.

There seems to be no related work on how to approach normalization of MQM results. In all the work published so far, authors simply count the number of MQM tags and stop there.
Our normalization approach is rather straightforward: instead of counting just error tags produced by each annotator, we count the tokens that these errors are assigned to. 

Once these counts are divided by the total number of tokens in the system's output, they provide a ratio of tokens with errors, as shown in Equation (\ref{eq:1}):

\begin{equation}
\label{eq:1}
error\: ratio=\frac{output\: tokens\: with\: errors}{total\: output\: tokens}
\end{equation}

Given that, according to this equation, the numerator counts words in the output that contain an error, the ratio is biased in favour of systems that produce shorter output. 
However, this is not a problem in our setup, as our taxonomy includes an \textit{Omission} error category. So if a word, segment, or phrase (or whatever the annotators deem as the basic unit) is was not translated from the source sentence, the target sentence is tagged with an \textit{Omission} error. While counting error tokens for our error ratio, we assume that 1 token was omitted for every omission error in the output, and so every omission error was given one phantom token to latch on to. This allows us to perform the calculations and prevents translations that lack some of the information of the source language sentence from having a low error rate.

The results of our error ratio calculations again show that the PBMT system has the largest error/token ratio (0.2633), while the factored system has a smaller ratio (0.212), and the NMT system has the smallest one (0.1277).
This is further backed up by a pairwise chi-squared ($\chi^2$) statistical significance test \citep{plackett1983karl}; we calculate statistical significance from 2x2 contingency tables for every system pair (PBMT x Factored, PBMT x NMT and Factored x NMT). 
In one such contingency table, the rows contain token counts for each of the systems, while the columns contain counts of tokens with and without errors. The null-hypothesis in this setting states that there is no link between the MT system and the number of tokens with or without errors that it produces (i.e. that no matter which system is employed, the number of errors is relatively similar). With the $p$ value lower than 0.0001 in all three comparisons, we can safely dismiss the null hypothesis, showing that the difference in the total counts of tokens with errors is statistically significant for all three system pairs. 

These error/token ratios provide an overall score for each system. At this point we would like to delve deeper and discover the performance of each system for each error type. To this end, we repeated these same measurements, but instead of performing them on all error types concatenated, they were performed separately for each specific error category.
The combined results of the aforementioned calculations and transformations are presented in Table \ref{t:MQM-token}.

\begin{table}[htbp]
\begin{center}
\begin{tabular}{lrr|rr|rr}
\hline
  & \multicolumn{2}{c|}{\bf PBMT}  & \multicolumn{2}{c|}{\bf Factored} & \multicolumn{2}{c}{\bf NMT} \\ 
\hline
\bf Error type & \bf OK & \bf Error & \bf OK & \bf Error & \bf OK & \bf Error \\
\hline
Accuracy & 3467 & 369 & \cellcolor{green!25}3525 & \cellcolor{green!25}291* & 3402 & 266 \\
\hspace*{1em}Mistranslation & 3547 & 289 & \cellcolor{green!25}3586 & \cellcolor{green!25}230* & 3471 & 197 \\
\hspace*{1em}Omission & 3801 & 35 & 3793 & 23 & \cellcolor{red!25}3619 & \cellcolor{red!25}49* \\
\hspace*{1em}Addition & 3814 & 22 & 3797 & 19 & 3655 & 13 \\
\hspace*{1em}Untranslated & 3813 & 23 & 3797 & 19 & \cellcolor{green!25}3662 & \cellcolor{green!25}6* \\
  &  &  &  &  &  & \\
Fluency & 3195 & 641 & \cellcolor{green!25}3298 & \cellcolor{green!25}518* & \cellcolor{green!25}3465 & \cellcolor{green!25}188** \\
\hspace*{1em}Unintelligible & 3790 & 46 & 3769 & 47 & \cellcolor{green!25}3668 & \cellcolor{green!25}0** \\
\hspace*{1em}Register & 3810 & 26 & 3794 & 22 & 3646 & 22 \\
\hspace*{1em}Spelling & 3833 & 3 & 3812 & 4 & 3659 & 9 \\
\hspace*{1em}Grammar & 3270 & 566 & \cellcolor{green!25}3371 & \cellcolor{green!25}445** & \cellcolor{green!25}3497 & \cellcolor{green!25}156** \\
\hspace*{2em}Word order & 3752 & 84 & 3752 & 64 & \cellcolor{green!25}3646 & \cellcolor{green!25}22** \\
\hspace*{2em}Function words & 3801 & 35 & 3780 & 36 & \cellcolor{green!25}3650 & \cellcolor{green!25}18* \\
\hspace*{3em}Extraneous & 3829 & 7 & 3810 & 6 & 3664 & 4 \\
\hspace*{3em}Incorrect & 3810 & 26 & 3790 & 26 & \cellcolor{green!25}3655 & \cellcolor{green!25}13* \\
\hspace*{3em}Missing & 3834 & 2 & 3812 & 4 & 3667 & 1 \\
\hspace*{2em}Word form & 3389 & 447 & \cellcolor{green!25}3471 & \cellcolor{green!25}345* & \cellcolor{green!25}3538 & \cellcolor{green!25}102** \\
\hspace*{3em}Part of speech & 3822 & 14 & 3800 & 16 & \cellcolor{green!25}3663 & \cellcolor{green!25}5* \\
\hspace*{3em}Tense... & 3775 & 61 & 3765 & 51 & \cellcolor{green!25}3648 & \cellcolor{green!25}20* \\
\hspace*{3em}Agreement & 3466 & 370 & \cellcolor{green!25}3540 & \cellcolor{green!25}276* & \cellcolor{green!25}3566 & \cellcolor{green!25}102** \\
\hspace*{4em}Number & 3778 & 58 & 3772 & 44 & \cellcolor{green!25}3646 & \cellcolor{green!25}22* \\
\hspace*{4em}Gender & 3788 & 48 & 3756 & 60 & \cellcolor{green!25}3644 & \cellcolor{green!25}24* \\
\hspace*{4em}Case & 3614 & 222 & \cellcolor{green!25}3694 & \cellcolor{green!25}122* & \cellcolor{green!25}3622 & \cellcolor{green!25}46** \\
\hspace*{4em}Person & 3836 & 0 & 3816 & 0 & 3664 & 4 \\
\hline
Total errors & 2826 & 1010 & \cellcolor{green!25}3007 & \cellcolor{green!25}809** & \cellcolor{green!25}3199 & \cellcolor{green!25}469** \\
\hline 
\end{tabular}
\caption{Processed annotation data from both annotators concatenated: each system's total number of tokens with and without errors. Statistical significance for a system, when compared to the system on its left, is marked with * where $p$-value is \textless0.05 and ** where $p$-value is \textless0.0001. Cells with a green background indicate that the system has fewer errors than the one on its left, while those in red indicate that it has more. In both cases, the green/red background is only displayed when the difference between the error ratios is statistically significant.
\label{t:MQM-token}}
\end{center}
\end{table}

We can derive several findings from this table.  
As mentioned earlier, looking at the grand total of tokens with and without errors, the difference between the systems is statistically significant by a wide margin. When looking at PBMT and factored PBMT, the factored system has significantly fewer errors than the pure PBMT system. The overall error rate is in this case reduced by 20\% (809 vs 1010 errors, cf. last row in Table \ref{t:MQM-token}). 
In addition, a separate analysis of specific error types that contribute to this score reveals that only some of the error categories are significantly different between the two systems. In the table, those categories are filled in with a green background. One can see that, when it comes to agreement errors, the only agreement error type that results in a significantly smaller number of errors with the factored PBMT system compared to the pure PBMT system is agreement in case.

However, taking a look at NMT shows that, not only does it result in a 42\% overall error reduction compared to the factored system (469 vs 809 errors), and 54\% with respect to pure PBMT (469 vs 1010 errors), but it also produces even less agreement errors -- overall, as well as at the level of number, gender and case -- while not using any kind of explicit linguistic information. This might in part be due to the use of sub-word segmentation, as inflections in Croatian are relatively regular. 
In addition to improving in the Agreement category, NMT also produces significantly fewer errors in many more categories than the factored model does.
Interestingly, it produces more Omission errors than either of the other two systems. It seems that NMT tends to sacrifice completeness of translation in order to increase overall fluency. This result is compatible with the average token per sentence ratio mentioned above: the NMT system has the lowest one (18.36; while PBMT has 18.99 and factored PBMT has 18.89).

\section{Additional Agreement Annotation}
\label{s:additionalagreement}

In this section we look at the agreement error category in more detail. 
Our motivation for picking this error type is twofold: (i) significant gains have been obtained in this error category (cf. Table \ref{t:MQM-token}) by NMT compared to the two PBMT systems, and (ii) this error category constitutes the main branch that we added to the core MQM tagset (to be able to evaluate the performance of MT on relevant linguistic phenomena present in Slavic languages, cf. Figure \ref{fig:slavic_set}).

Agreement is also worth exploring further because two syntactically different types of agreement are subsumed under the MQM Agreement tags, namely:

\begin{itemize}
\item Local, short-distance agreement (or phrase agreement), which concerns agreement of elements within a phrase.\footnote{Unlike in SMT jargon, here a phrase refers to a grammatical unit, not just a string of contiguous words.}
\item Long-distance agreement (or sentence agreement), which concerns agreement of elements at the sentence level, outside phrase boundaries. These elements have wider spans and can be much further apart.
\end{itemize}

For example, local agreement would be agreement between an adjective and a noun, or between a preposition and the following noun, while sentence agreement would be agreement between a noun and a verb.
Table \ref{e:agr} contains an example of agreement errors at these two levels. The phrase bolded in the first sentence contains disagreement in case: the preposition ``u" should introduce a phrase in the dative case (``palijativn\textbf{oj} skrbi"), but the translation is in the accusative case (``palijativn\textbf{e} skrbi"), which is morphologically marked. The phrase bolded in the second sentence contains disagreement in gender: the noun ``jedinica" (``unit", feminine) is the subject of the sentence and as such should agree with the verb ``nastati" (``was created") that follows it in gender, number, case and person; however, in the translation, the verb is marked for masculine gender (``nasta\textbf{o}") instead of the required feminine (``nasta\textbf{la}").

\begin{table}[htbp]
\begin{center}
\begin{tabular}{ll}
Phrase disagreement: & Veliki broj ljudi radi \textbf{u palijativne skrbi}. \\
Sentence disagreement: & Stalna antikorupcijska \textbf{jedinica}, koja se bori protiv \\ 
& svakog oblika korupcije, \textbf{nastao} je 2011. godine. \\
\end{tabular}
\caption{Example sentences showcasing the two different spans an agreement error can take. The first sentence features disagreement in case, whereas the second one features disagreement in gender.
\label{e:agr}}
\end{center}
\end{table}

This distinction is important not only linguistically, but can also  be informative from a technical perspective. Thus, we conducted an additional layer of annotation outside the framework of MQM: each agreement error was categorized as corresponding to either phrase or sentence level. 
Additionally, the type of elements participating in the error was marked as well, in order to obtain more fine-grained insights. 

For phrase agreement, the phrases in question can be prepositional phrases (PP) that contain a noun phrase (NP), noun phrases that contain an adjective (ADJ) and a noun (N), noun phrases comprised of two nouns (N+N) and noun phrases containing numerals (NUM+NP). In sentence agreement, elements that often need to agree are subjects and verbs (S+V, usually noun and verb), verbs and objects (V+O, usually verb and noun), two or more noun phrases coordinated with a conjunction (NP+C+NP, usually ``i" [``and"]), and a noun phrase followed by a subordinating conjunction (NP+CSUB, usually ``koji/koja/koje" [``which" or ``that"]).
The results of applying this categorisation to our dataset are presented in Table \ref{t:agreement}.

\begin{table}[htbp]
\begin{center}
\begin{tabular}{lrrr|lrrr}
\hline
  & \multicolumn{3}{c|}{\bf Phrase agreement}  & & \multicolumn{3}{c}{\bf Sentence agreement}\\ 
\hline
Elements & \bf PBMT & \bf Factored & \bf NMT & Elements & \bf PBMT & \bf Factored & \bf NMT \\
\hline 
PP+NP & 24 & 14 & 3 & S+V & 20 & 19 & 7 \\
ADJ+N & 15 & 5 & 2 & V+O & 5 & 3 & 3 \\
N+N & 4 & 7 & 1 & NP+C+NP & 6 & 7 & 1 \\ 
NUM+NP & 1 & 1 & 0 & NP+CSUB & 1 & 2 & 0 \\
\hline 
\bf Total & \bf 44 & \bf 27 & \bf 6 & \bf Total & \bf 32 & \bf 31 & \bf 11 \\
\hline 
\end{tabular}
\caption{Breakdown and categorization of agreement errors found in the annotated data.
\label{t:agreement}}
\end{center}
\end{table}

As the table shows, the factored PBMT model leads to quite a large improvement upon pure PBMT when it comes to phrase agreement, but the improvement is almost negligible when it comes to sentence agreement (phrase agreement sees a $\sim$38\% relative reduction in errors, while the number of sentence agreement errors is reduced by $\sim$4\% relative). Meanwhile, the NMT model produces substantially less agreement errors of both agreement types ($\sim$86\% relative reduction in phrase agreement errors and $\sim$66\% relative reduction in sentence agreement errors, when compared to pure PBMT). 

Knowing that both the factored model and NMT model produce less agreement errors overall when compared to PBMT (cf. Table \ref{t:MQM-token}), it is no surprise that they produce overall less of either level (phrase and sentence) of agreement errors.
However, just as in the MQM analysis conducted in the previous section, simply counting errors is not enough to know whether the difference in the number of errors between two MT paradigms is statistically significant.
Thus, to determine whether these differences are statistically significant overall, we once again normalized the errors to the token level and employed a chi-squared ($\chi^2$) test.
We calculate statistical significance from 2x2 contingency tables for every system pair (PBMT x Factored, PBMT x NMT and Factored x NMT), for each type of error (overall phrase agreement and overall sentence agreement), as well as for the elements that make up these errors.
In these contingency tables, rows contain token counts for each system, while columns contain counts of tokens with and without agreement errors. The null-hypothesis states that there is no link between the MT system and the frequency of a given agreement error that it produces.

\begin{table}[htbp]
\begin{center}
\begin{tabular}{lrr|rr|rr}
\hline
  & \multicolumn{2}{c|}{\bf PBMT}  & \multicolumn{2}{c|}{\bf Factored} & \multicolumn{2}{c}{\bf NMT} \\ 
\hline
\bf Error type & \bf OK & \bf Error & \bf OK & \bf Error & \bf OK & \bf Error \\
\hline
\bf Total & & & & & & \\
\hspace*{1em}Phrase & 1811 & 88 & \cellcolor{green!25}1835 & \cellcolor{green!25}54* & \cellcolor{green!25}1824 & \cellcolor{green!25}12** \\
\hspace*{1em}Sentence & 1835 & 64 & 1827 & 62 & \cellcolor{green!25}1814 & \cellcolor{green!25}22** \\
\hline
\bf Phrase agreement & & & & & & \\
\hspace*{1em}PP+NP & 1851 & 48 & \cellcolor{green!25}1861 & \cellcolor{green!25}28* & \cellcolor{green!25}1830 & \cellcolor{green!25}6* \\
\hspace*{1em}ADJ+N & 1869 & 30 & \cellcolor{green!25}1879 & \cellcolor{green!25}10* & 1832 & 4 \\
\hspace*{1em}N+N & 1891 & 8 & 1875 & 14 & \cellcolor{green!25}1834 & \cellcolor{green!25}2* \\
\hspace*{1em}NUM+NP & 1897 & 2 & 1887 & 2 & 1836 & 0 \\
\hline
\bf Sentence agreement & & & & & & \\
\hspace*{1em}S+V & 1859	& 40 & 1851 & 38 & \cellcolor{green!25}1822 & \cellcolor{green!25}14* \\
\hspace*{1em}V+O & 1889	& 10 & 1883 & 6 & 1830 & 6 \\
\hspace*{1em}NP+C+NP & 1887 & 12 & 1875 & 14 & \cellcolor{green!25}1834 & \cellcolor{green!25}2* \\
\hspace*{1em}NP+CSUB & 1897	& 2 & 1885 & 4 & 1836 & 0 \\
\hline
\end{tabular}
\caption{Normalized agreement annotation data: each system's total number of tokens with and without agreement errors, also including data with regards to which elements contained errors. Statistical significance for a system, when compared to the the one on its left, is marked in green. If $p$-value is \textless0.05, it is marked with *, and ** where $p$-value is \textless0.0001.
\label{t:agreement-token}}
\end{center}
\end{table}

As shown in Table \ref{t:agreement-token}, the total counts show that when looking at phrase agreement, there is steady improvement between the systems: the factored system has significantly less tokens with a phrase-agreement error than the PBMT system ($p$=0.004), while the NMT system has significantly less than the factored system does ($p$\textless0.0001). On the other hand, looking at sentence agreement and comparing pure PBMT to the factored PBMT model yields a $p$-value of 0.8799, revealing no statistical significance, while comparing the factored model to the neural model yields a $p$-value of 0.00002, indicating a statistically significant difference in the number of tokens with errors. In other words, when compared to PBMT, both the factored model and the NMT model significantly reduce the number of phrase-agreement errors, whereas the factored model does not significantly reduce the number of sentence-agreement errors, but the neural system does.

These results are in line with previous research that showed how, for the English-to-Croatian language pair, factored PBMT struggles with sentence agreement due to the limitations of n-gram language models: \citet{sanchez2016dealing} showed that using high-order language models (with order higher than $3$) for morphosyntactic tags leads to a degradation in translation quality because of the free word order of Croatian. On the contrary, the power of recurrent neural network units to model long-distance phenomena allows the NMT system to improve on both phrase and sentence agreement.

\section{Conclusion}\label{s:con}

This paper describes a fine-grained human evaluation of three approaches to MT (pure PBMT, factored PBMT and NMT). Our analysis has provided answers to several questions, one of which was the main drive behind the development of a factored system for English-to-Croatian: is there a way to better handle agreement when translating to a morphologically rich language?
We can now confidently claim that factored models result in significantly less agreement errors overall compared to pure PBMT, when translating from English to Croatian.

We can also confidently conclude that NMT handles all types of agreement better than both pure PBMT and factored PBMT, which corroborates the findings of other researchers' NMT evaluations conducted for other language pairs. Our NMT system produces sentences with far fewer errors, and output that is more fluent and more grammatical, which should be of help when it comes to the task of post-editing. 

Furthermore, the error taxonomy that was developed for this research, while only used for the English-to-Croatian language direction in the current work, should be applicable for the analysis of errors for any translation direction towards a Slavic language, as it takes into account specific grammatical properties shared by the members of this language family.

Among other possible lines of future work, including the application of our methodology to another language pair that involves a Slavic target language (e.g. English--Czech), performing more controlled IAA analysis or IAA adjudication, as well as comparing to an NMT model without sub-word segmentation, another direction to go in is further adapting the tagset. In its current version, it has been demonstrated to be informative when comparing PBMT to factored PBMT. However, NMT has shown itself to produce language that is so fluent that the fine-grained hierarchy in the \textit{Fluency} branch is of little use. Meanwhile, the most common error type in the NMT output is \textit{Mistranslation}, which, according to the MQM guidelines, covers both lexical selection and (less intuitively) translation of grammatical properties (e.g. if `cats[pl.]' is translated into Croatian as `ma\v{c}ka[sg.]', this is to be tagged as \textit{Mistranslation}, in spite of correct lexical choice). This makes it quite a vague category, so if one would wish to perform an even more nuanced analysis of errors for NMT, adding additional layers to the \textit{Accuracy} branch would seem a promising direction to follow.

\begin{acknowledgements}

We would like to extend our thanks to Maja Popovi\'{c}, who provided invaluable advice, and Denis Kranj\v{c}i\'{c}, who performed the annotation together with Filip Klubi\v{c}ka, first author of the paper. This research was partly funded by the ADAPT Centre, which is funded under the SFI Research Centres Programme (Grant 13/RC/2106) and is co-funded under the European Regional Development Fund. This research has also received funding from the European Union Seventh Framework Programme FP7/2007-2013 under grant agreement PIAP-GA-2012-324414 (Abu-MaTran) and the Swiss National Science Foundation grant 74Z0\_160501 (ReLDI). 

\end{acknowledgements}

\bibliographystyle{spbasic}      % basic style, author-year citations

\bibliography{mybib}

\end{document}